\documentclass{IEEEtran}
\usepackage{amsmath,amsthm,amsfonts,amssymb,bm,mathrsfs}
\usepackage{subfigure}
\usepackage{graphicx,multicol}
\usepackage{algorithm}
\usepackage{algorithmic}
\usepackage{multirow,booktabs,threeparttable}
\usepackage{color,cite,url}

\usepackage{setspace}
\theoremstyle{remark}

\theoremstyle{plain}

\begin{document}
\title{Internet of Intelligence: The Collective Advantage for Advancing Communications and Intelligence}

\author{
Rongpeng Li, Zhifeng Zhao, Xing Xu, Fei Ni, and Honggang Zhang

\thanks{R. Li, X. Xu, F. Ni, and H. Zhang are with Zhejiang University, Hangzhou 310027, China, (email: \{lirongpeng, hsuxing, 11931047, honggangzhang\}@zju.edu.cn).}

\thanks{Z. Zhao is with Zhejaing Lab as well as Zhejiang University, Hangzhou 310012, China (email: zhaozf@zhejianglab.com).}

\thanks{This work was supported in part by National Key R\&D Program of China (No. 2017YFB1301003), National Natural Science Foundation of China (No. 61701439, 61731002), Zhejiang Key R\&D Program of China (No. 2019C01002, 2019C03131).}
}

\maketitle

\begin{abstract}
The fifth-generation cellular networks (5G) has boosted the unprecedented convergence between the information world and physical world. On the other hand, empowered with the enormous amount of data and information, artificial intelligence (AI) has been universally applied and pervasive AI is believed to be an integral part of the six-generation cellular networks (6G). Consequently, benefiting from the advancement in communication technology and AI, we boldly argue that the conditions for collective intelligence (CI) will be mature in the 6G era and CI will emerge among the widely connected beings and things. Afterwards, we highlight the potential huge impact of CI on both communications and intelligence. In particular, we
introduce a regular language (i.e., the information economy metalanguage) supporting the future collective communications to augment human intelligence and explain its potential applications in naming Internet information and pushing information centric networks forward. Meanwhile, we propose a stigmergy-based federated collective intelligence and demonstrate its achievement in a simulated scenario where the agents collectively work together to form a pattern through simple indirect communications. In a word, CI could advance both communications and intelligence.
\end{abstract}

\section{Introduction}
In the year of 2019, the fifth-generation cellular networks (5G) has come into the commercialization phase. With the support of enhanced mobile broadband (eMBB) service, massive-connected machine-type communication (mMTC) service and ultra-reliable low-latency (URLLC) service, 5G is expected to transform cellular network from pure life-based Internet to industry-oriented Internet. 5G begins to classify services into three types (e.g., eMBB, URLLC, mMTC) and separately provision each service using one infrastructure, by leveraging the softwarization and virtualization of network functionalities as well as network slicing techniques to appropriately orchestrate resources. Besides, 5G has introduced the network data analytics function (NWDAF) to better exploit the availability of massive actionable data. In the physical layer, 5G accommodates stringent requirements in terms of data rate and latency, by introducing millimeter-wave (mmW) communications, exploiting massive multiple-input/multiple-output (MIMO) links, and deploying (ultra) dense radio access points \cite{calvanese_strinati_6g_2019}. Therefore, the transmission rate is speculated to be significantly improved in 5G with further reduced latency.

\begin{figure}
	\centering
	\includegraphics[width=0.475\textwidth]{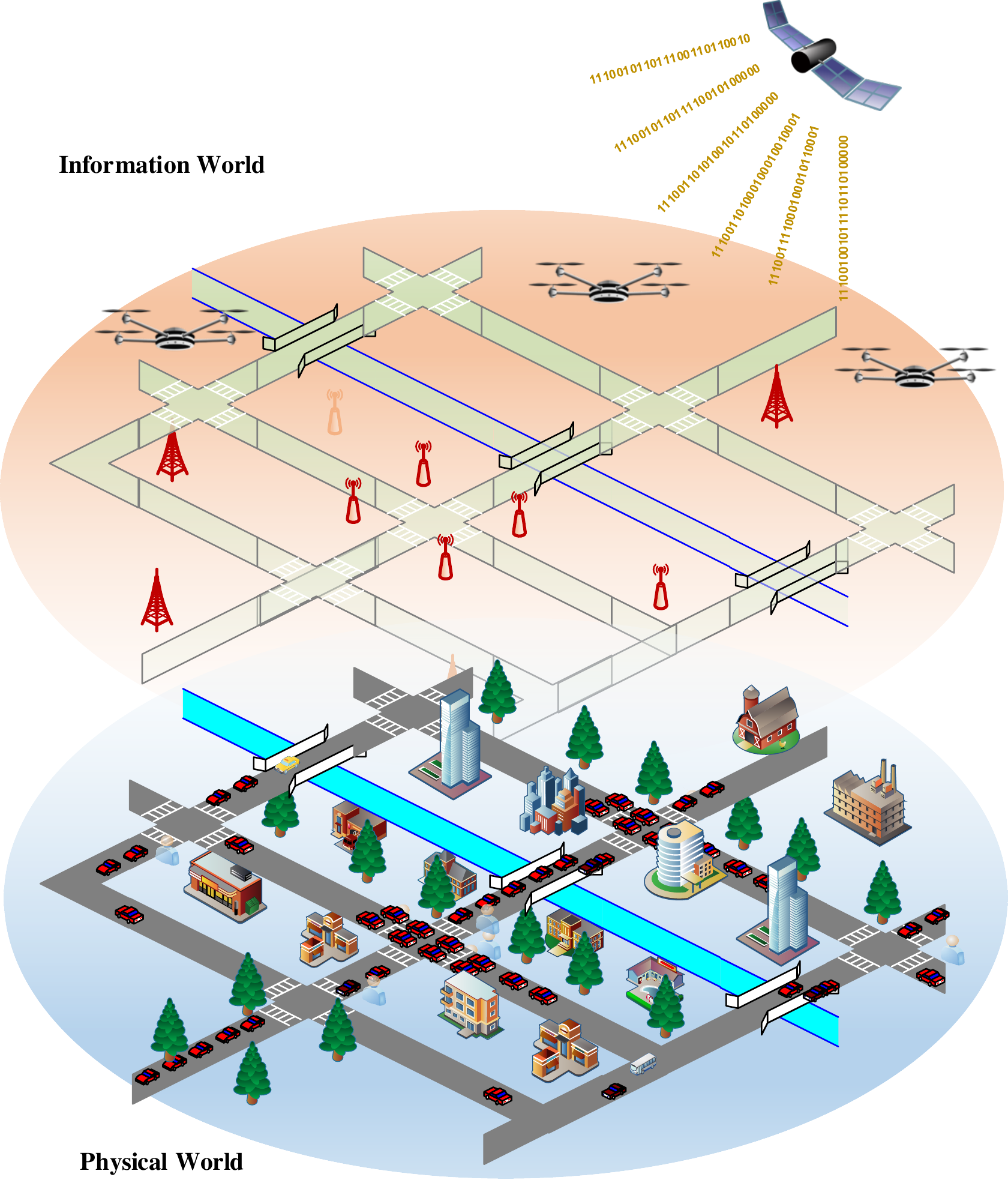}
	\caption{An illustration of the unprecedented convergence between information world and physical world.}
	\label{fig:Future_Info_Physical_Conver}
\end{figure}

Meanwhile, researchers from both academia and industry have been actively looking forward to techniques for the six-generation cellular networks (6G)\footnote{Notably, some in the literature readily used the term 6G while others use terminologies like beyond 5G (B5G) or 5G+. In this article, we assume these terminologies are inter-changeable.}. Despite somewhat unclear research directions (e.g., tera-Hertz communications, space-air-ground integrated network \cite{niu_space-air-ground_2017}, etc), 6G has been destined to further bring immerse Internet experience. Since artificial intelligence (AI) manifested the astonishing capability in the well-known success of AlphaGo \cite{silver_mastering_2016}, there is little doubt that everything around us will be very intelligent with the enormous amount of data and information, and pervasive AI will be an integral part of 6G as well \cite{li_intelligent_2017,gerla_internet_2014,saad_vision_2019,calvanese_strinati_6g_2019}, so as to reap the collected data in NWDAF. Following this trend, 6G contributes to an unprecedented convergence between the physical world and the information world by bringing benefits to cutting-edge technologies such as the digital twin. Moreover, the network architecture in the 6G era could be revolutionized and emerge in a more hierarchical and de-centralized manner. In other words, as illustrated in Fig. \ref{fig:Future_Info_Physical_Conver}, rather than the completely base station (BS)-centric networking infrastructure in 5G and pre-5G, 6G will more aggressively support the direct communications among vehicles and things, assisted by satellites, hovering UAVs/ballons and ground BSs. 

One field that will first enjoy the advancement in communications technology (CT) might be the Internet of vehicles. Recently, the weird phenomena where the Internet of vehicles only implies a collection of sensor platforms that provide information to drivers and upload filtered sensor data (e.g., GPS location, road conditions, etc.) to the cloud, has been experiencing remarkable changes. As the Level-4 autonomous car soon becomes a reality\footnote{\url{https://www.engadget.com/2018/11/02/volvo-baidu-autonomous-cars-china/}}, a fleet of vehicles in the future needs to exchange their sensor inputs among each other so as to realize prompt delivery of passengers to destination with maximum comfort and minimum impact on the surrounding environment. The vehicle to everything (V2X) turns to rely on a combination of eMBB, mMTC and URLLC services rather than the traditionally simple sensor-data acquisition \cite{gerla_internet_2014}. In particular, V2X connects vehicles with other vehicles (i.e., V2V), people with wearable devices or smartphones (i.e., V2P), infrastructure (i.e., V2I) as well as the network (i.e., V2N). Meanwhile, AI will play a crucial role in this collective V2X system, so as to avoid inter-vehicle collision, protect people nearby and realize smart traffic control. Consequently, V2X could enjoy the advancement in both CT and AI and possibly produce a collective intelligence (CI). 

Indeed, as a non-nascent concept, CI is a form of ``universally distributed intelligence, constantly enhanced, coordinated in real time, and resulting in the effective mobilization of skills" \cite{levy_semantic_2011}. Therefore, being significantly different from swarm intelligence, which is inspired from the collective behavior in biology and usually refers to a general set of centralized algorithms (e.g., ant colony optimization), the goal of CI is ``mutual recognition and enrichment of individuals rather than the cult of fetishized or hypostatized communities" \cite{levy_semantic_2011}. CI has been successfully applied in several well-known examples including Wikipedia and reCaptcha, by enabling humans to interact and to share and collaborate with both ease and speed \cite{flew_new_2008}. For Wikipedia, the Internet gives participating humans the opportunity and privilege to store and to retrieve knowledge through the collective access to these databases \cite{flew_new_2008}. With the development of the Internet and its widespread use, the opportunity to contribute to knowledge-building communities, is greater than ever before. Meanwhile, as discussed above, increasing numbers of things will be surely connected in a more efficient and reliable manner in the 6G era. Therefore, we boldly argue that CI will enter into a more advanced level (i.e., Internet of intelligence, IoI) and drive the digitization of information \& communication. 

In this article, we present the positive impact of CI on communications and intelligence from two aspects:
\begin{itemize}
	\item \emph{Advancing Communications:} Currently, both information-centric networks and mobile-edge computing allow in-way routers to cache some popular contents, so as to shorten the end-to-end latency. In Section \ref{sec:adv_com}, we boldly argue such a caching operation could be more aggressive by adopting a reasoning language (i.e., the information economy metalanguage (IEML) \cite{levy_semantic_2011}) in routers and explain the fundamental design principles of IEML. 
	\item \emph{Advancing Intelligence:} The reliable communication with rather low latency lays the very foundation for distributed agents at different locations or positions to coordinate and cooperate. In Section \ref{sec:adv_int}, we talk about a CI-inspired method and demonstrate the impact of communication links on the inter-agent coordination. 
\end{itemize}

\section{CI: Advancing Communications}
\label{sec:adv_com}
\subsection{The Communicating Language among Collective Intelligent Agents}

In order to fulfill the goal of openness and sharing in CI as well as to effectively augment human intelligence, it is incentive to design a communicating language among the agents. As human intelligence is intimately tied to the capacity to understand linguistic expressions and manipulate symbolic expressions \cite{levy_semantic_2011}, a scientific model of language with computable semantics is highly required. We strongly argue that IEML \cite{levy_semantic_2011} is an appropriate candidate to achieve such a goal.

\begin{figure*}
	\centering
	\includegraphics[width=0.85\textwidth]{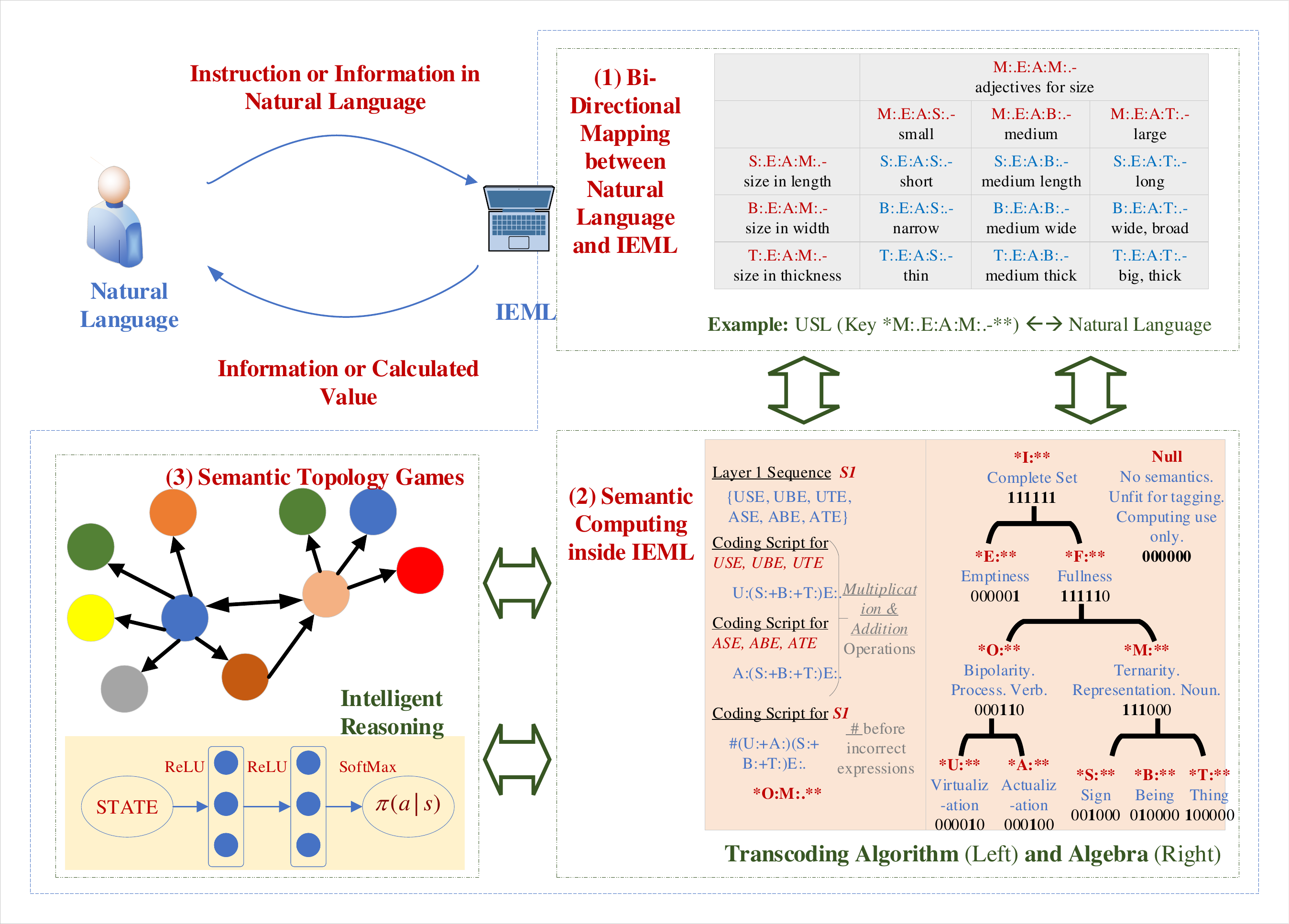}
	\caption{The basics of IEML, a regular language for the CI.}
	\label{fig:IEML}
\end{figure*}

As a regular language, IEML adopts an alphabet with six elementary symbols $\{E, U, A, S, B, T\}$. The lower-right part of Fig. \ref{fig:IEML} highlights the six elementary symbols and explains its related meaning in natural language. Basically, IEML builds semantic sequences by concatenating symbols \cite{levy_semantic_2011}. If the length of one sequence is denoted as the number of elementary symbols therein, the sequences of IEML can have only 7 predefined lengths, namely $1$, $3$, $9$, $27$, $81$, $243$, $729$ from Layer $0$ to Layer $6$. Terminating punctuation marks (i.e.,  ``:", ``.", ``{}-{}", ``'", ``,", ``\_", ``;") are used to express the end of sequences from Layer $0$ to Layer $6$. IEML takes advantage of a script to reciprocally transcode the algebra into sequences. The various algebraic expressions (e.g., $\{OSE\}$ and $\{USE, ASE\}$), which are equivalent in terms of meaning in the real world, are translated by a unique code in the script, and vice versa. During the translation, two operations (i.e., multiplication and addition) are defined. Different from the common meaning in mathematics, the multiplication operation denotes a reversible means to construct sequences for a particular layer from the sequences from its lower layer. Meanwhile, the addition operation means to manipulate sets of sequences by union, intersection or symmetrical difference. Furthermore, IEML produces and recognizes words, phrases, texts and hyper-texts by the IEML dictionary, which is composed of a set of inter-operable keys. The upper-right part of Fig. \ref{fig:IEML} shows an example to interpret the key \text{*M:.E:A:M:.-**}. As the substance of \text{*M:.E:A:M:.-**} is \text{*M:.**} (i.e., a noun), \text{*M:.E:A:M:.-**} could be regarded as the adjectives for size. Specifically, the substance of \text{*S:.E:A:T:.-**} is \text{*S:.**}, which corresponds to the length of objects, and \text{*E:A:T:.**} implies large quantities \cite{levy_semantic_2011} as its attribute, we can infer \text{*S:.E:A:T:.-**} means ``long''. Similarly, it can be observed from Fig. \ref{fig:IEML}, all the adjectives for size can be found in \text{*M:.E:A:M:.-**}.


As a regular language, IEML has the following merits.
\begin{itemize}
\item The meaning of texts can be analyzed and represented by uniform semantic locators (USLs, i.e., sets of IEML sequences) and their graphs of relationships, which are represented in semantic variables by transforming between concepts and texts. Moreover, all operations based on transformations of semantic variables in IEML are computable by following the properties of finite state machines \cite{levy_semantic_2011}. Therefore, IEML is endowed with the inborn computing capability and facilitates the building of knowledge graphs together with some artificial intelligence algorithms as depicted in the lower-left part of Fig. \ref{fig:IEML}. 
\item Instead of being mutually exclusive to other high-level programming languages (e.g., XML (Extensible Markup Language), RDF (Resource Description Framework) and OWL (Ontology Web Language)), IEML could be written in these programming languages to generate and handle texts \cite{levy_semantic_2011}. Therefore, IEML paves the way to perform trans-linguistic hyper-textual communication 
after the careful design of an open and universal encyclopedic library, so as to enable creative conversations among collective intelligent agents. Hence, it becomes more flexible to build a knowledge graph with much more abundant information across languages.
\end{itemize}

\subsection{IEML: Pushing Information-Centric Networks Forward}

\begin{figure*}
	\centering
	\includegraphics[width=0.825\textwidth]{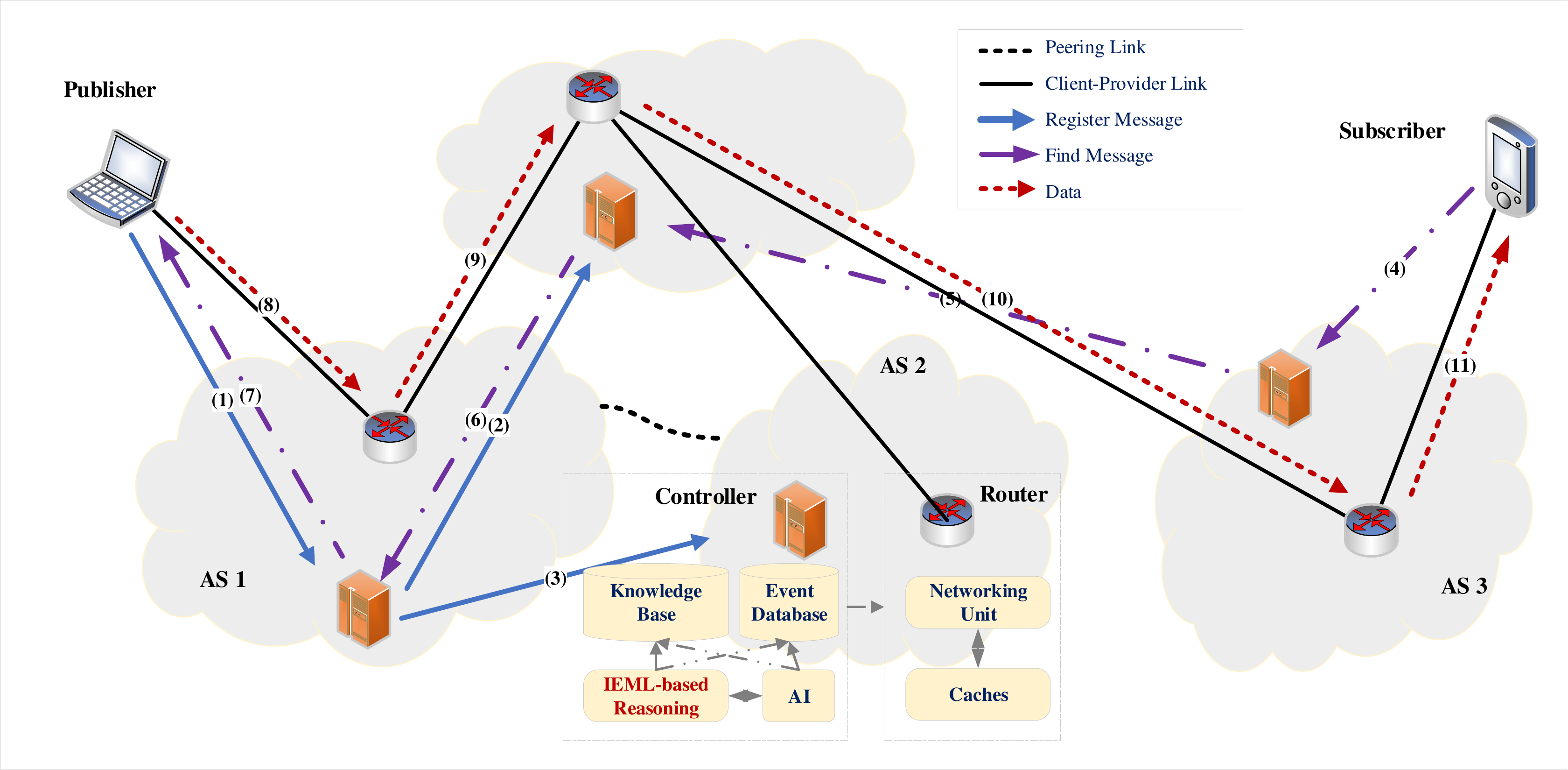}
	\caption{An illustration of IEML-based ICN Networks.}
	\label{fig:IEML_icn}
\end{figure*}

As 5G has connected beings and things in a unprecedented manner, the Internet has shifted from traditional pair-wise communications between end-hosts to information dissemination. Consequently, information-centric networking (ICN) has
emerged as a promising candidate for the architecture of
the future Internet by naming information at the network layer and deploying in-network caching \cite{xylomenos_survey_2014}. Besides, ICN could realize information-awareness as well as mobility management and security enforcement via information-naming related regulations. 

Fig. \ref{fig:IEML_icn} gives an illustrated IEML-based ICN architecture inspired by the data oriented network architecture
(DONA) \cite{koponen_data-oriented_2007}. In particular, the architecture shares many similarities from DONA. For example, it replaces the hierarchical uniform resource locators (URLs) with IEML-based names. Therefore, since the information is decoupled from locations and could be cached in-path, the support for information mobility and availability is significantly enhanced. Besides, similar to DONA, a name resolution mechanism will be deployed as an overlay to maintain traditional IP addressing and routing. But, the only yet major difference from DONA is that the proposed architecture adopts an IEML-based naming method.

The argument to introduce IEML to ICN is not trivial, since the naming of information objects and services is one of the fundamental design choices in each ICN architecture and shapes all other aspects of the architecture \cite{xylomenos_survey_2014}. Basically, there exist two types of naming methods for ICN, that is, hierarchical naming and flat naming. Consistent with the traditional hierarchical URLs, hierarchical names facilitate the information aggregation and benefit the Internet with over tens of trillions of named information objects. Typically,  hierarchical names imply to allocate names with the same prefix to point at the same area. Though this location-binding solution makes it easier for human to read and understand, it is also identified as one of the main shortcomings of the current Internet architecture with respect to mobility support \cite{xylomenos_survey_2014}. On the other hand, it is not easy to directly aggregate traditional flat names and huge routing and/or name resolution tables often become a necessity. In addition, it is required to update these data structures whenever the information moves. Fortunately, IEML could compensate the shortcomings of traditional flat names as it is native for self-reasoning and information aggregation. Besides, since IEML and human languages are mutually interpretable, it is competent to understand IEML. Furthermore, consistent with other flat names, IEML avoids the location-identity binding (which is necessary for hierarchical names), thus simplifying mobility. Moreover, IEML brings some additional advantages like enhanced content caching. For example, the computable advantage of transformations of semantic languages in IEML provides a deeper understanding of information and its combination with AI-based popularity prediction methods facilitates routers to better infer some popular contents. Therefore, we boldly argue that IEML could help to push ICN forward.

\section{CI: Advancing Intelligence}
\label{sec:adv_int}

\subsection{Stigmergy-Enhanced Federated Collective Intelligence}

\begin{figure*}
	\centering
	\includegraphics[width=0.745\textwidth]{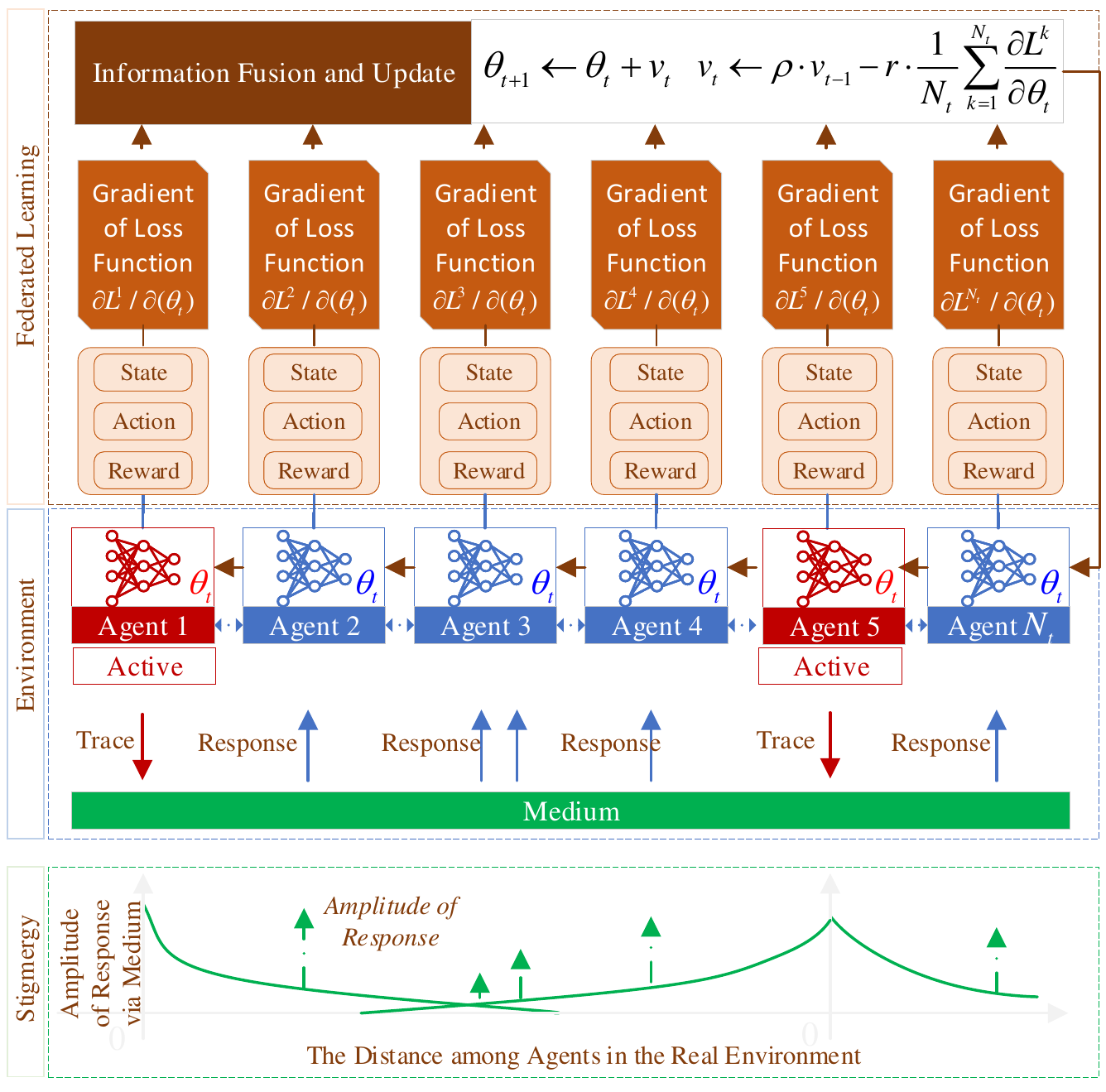}
	\caption{An illustration of stigmergy-enhanced federated CI (SEAL) with multiple active and inactive agents. The amplitude of response received by inactive agents from active one are distance-based. The neural networks within each agent is updated in a federated manner by collecting the gradient of loss function perceived by each agent.}
	\label{fig:stigmergy}
\end{figure*}

V2X is one of most promising scenarios to extend the realm of social world to include multiple autonomous cars collectively committing to one goal. Typically, compared with single agent's intelligence, such a collective interaction among multiple agents implies enhanced robustness, improved scalability, and reduced cost \cite{stone_multiagent_2000}. For example, limited by the size and power of devices, a single agent often fails to fulfill computation-heavy tasks and the coordination among multiple agents becomes indispensible.

Recently, single-agent machine learning has manifested the intelligence in astonishing achievements (e.g., AlphaGo powered by deep reinforcement learning). However, the extension of reinforcement learning from single-agent scenarios to multi-agent scenarios (MAS) is not straightforward, since an agent in MAS normally observes the global environment partially and the inclusion of other agents often makes learning environment non-stationary. Hence, an effective communication method for information exchange and solution synthesis is necessary for the coordination and cooperation in MAS. 

In this article, we present a Stigmergy-Enhanced federAted collective inteLligence (SEAL) algorithm for MAS. As its name implies, this SEAL algorithm consists of two fundamental ingredients, that is, stigmergy and federated learning. Originating from entomology, the concept of ``stigmergy" has been inspired by the behaviors of social insects and first introduced by French entomologist Pierre-Paul Grass\`e in 1950s \cite{hsu_brain-inspired_2019}. Based on the ``medium'', an information aggregator in the framework of stigmergy, multiple agents can interact with each other indirectly to reduce their behavioral localities. On the other hand, 
compared to centralized training approaches, federated learning is a de-centralized training approach which enables to collaboratively learn a machine learning model while keeping all the data with possibly private information locally. In such a case, agents can benefit from obtaining a well-trained machine learning model without compromising their privacy. 

Fig. \ref{fig:stigmergy} gives an illustration of the SEAL algorithm. Taking the scenario example of V2X, vehicles with on-board units could act as intelligent agents in Fig. \ref{fig:stigmergy} and communicate with each other directly (e.g., V2V) or indirectly via the medium (e.g., V2I). Considering that the movement of any vehicle certainly affects others' movement, the SEAL algorithm formulates this fact as that some agents receive a response or reward following other active agents' trace induced by certain actions. In particular, the response function could be a distance-dependent function like a Gaussian distributed function or other brain-inspired functions \cite{hsu_brain-inspired_2019}. On top of the communication mechanism, each agent will update its knowledge database (e.g., state, action, reward) and compute the gradient of the loss function with respect to the neural network parameters. Moreover, the existence of V2N facilitates the governance of vehicles by aggregating the knowledge from individual models. Therefore, consistent with the federated learning, the SEAL algorithm adopts a virtual agent in the cloud to fuse these gradients and train the neural network parameters.

\subsection{The Achievement by Collective Intelligent Agents}
\begin{figure*}
	\centering
	\includegraphics[width = 0.85\textwidth]{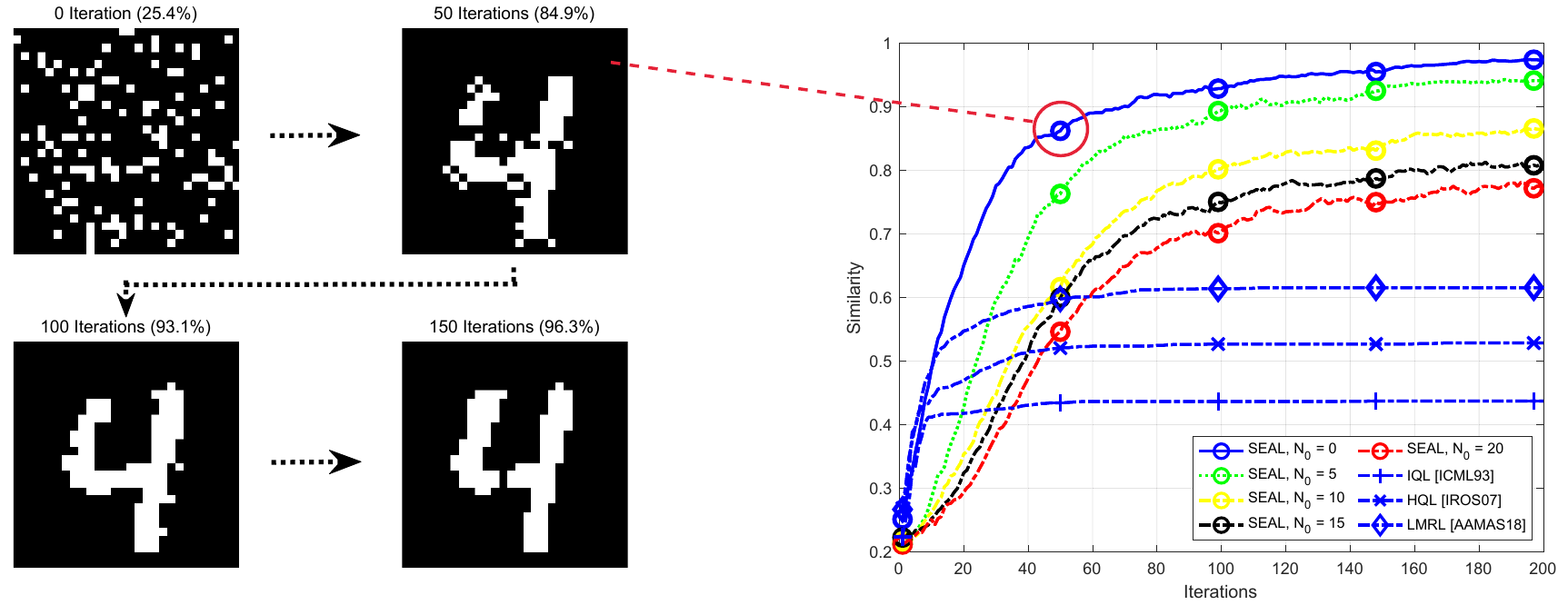}
	\caption{The similarity between the original image and the shape formed by collective intelligent agents via the indirect communication with different noise. Some typical algorithms are also included like IQL (independent $\mathcal{Q}$-learning) by M. Tan in ICML'93, HQL (hysteretic $\mathcal{Q}$-learning) by L. Matignon \textit{et al.} in IROS'07 and LMRL (lenient multi-agent reinforcement learning) by G. Palmer \textit{et al.} in AAMAS'18.}
	\label{fig:achievement}
\end{figure*}

In this part, we talk about some simulation results to demonstrate the achievement of SEAL. Concretely speaking, we assume there exists an area with some movable agents and their goal is to form a specific pattern of ``$4$" as depicted in Fig. \ref{fig:achievement}. Moreover, the total simulation area can be classified into the labeled area and the unlabeled area, and the labeled area corresponds to the specified shape which agents need to form. For example, in Fig. \ref{fig:achievement}, the total area is occupied by 28$\times$28 blocks while the labeled area colored in white needs 119 agents to fill it up. Consistent with Fig. \ref{fig:stigmergy}, each movement of the agents will leave some trace (i.e., the digital pheromone) as the medium of the indirect communications among agents while each block in the total area can keep the digital pheromone for a period of time in order to provide guidance for the cooperation of agents.	As for the movement of each agent, we set the following rules.
\begin{itemize}
	\item A certain proportion of agents are selected out for moving opportunities. Each active agent is supposed to move a block towards one of four directions at a time, that is, UP, DOWN, LEFT, and RIGHT.
	\item Each block in the area can only be occupied simultaneously by an agent.
	\item Each active agent can leave the digital pheromone after arriving a new position and sense the amount of digital pheromone within a certain range.
	\item Each agent can identify whether the current position is labeled or unlabeled.
	\item Each agent can sense the existence of neighbors in all directions (i.e., UP, DOWN, LEFT, and RIGHT) and can even communicate with them briefly.
\end{itemize}

We have applied the SEAL algorithm by classifying the cooperation between agents into the following five steps:
\begin{enumerate}
	\item \emph{Selecting active agents}: The possibility of each agent being selected is related to the priority of its action itself. Concretely, the agent only with the highest action priority can be regarded as ``active" and obtain the moving opportunity. The selection of active agents is also consistent with our intuition that common heterogeneous agents could respond to one task in terms of speed and delay distinctly.
	\item \emph{Selecting attractor}: Each block with a certain amount of digital pheromone can be regarded as an attractor. In this part, each active agent selects its own attractor in a stochastic manner. The probability of each attractor being selected corresponds to the amount of digital pheromone as well as the distance with the agent.
	\item \emph{Moving}: After active agents and attractors being determined, each active agent will move one block towards its own attractor. But this agent will stop or select another direction if it encounters the obstruction of neighbors. For example, an agent will move a block towards the ``UP" direction if the selected attractor is above this agent. However, if the block to be visited has already been occupied by another agent, this agent will stop or move towards the ``LEFT" or ``RIGHT" direction.
	\item \emph{Leaving digital pheromone}: When an agent arrives at a new position, it will leave the digital pheromone in the current position according to the attribute (e.g., labeled or unlabeled) of the area. Specifically, the amount of digital pheromone in this position will be increased if this position belongs to the labeled area; meanwhile it will be decreased otherwise. 
	\item \emph{Updating priority}: The action priority for a particular agent will be updated according to the information of its position as well as the number of surrounding neighbors. During the simulation, a predefined reward table is provided for the updating of action priority.	
\end{enumerate}

Fig. \ref{fig:achievement} provides the simulation results for this typical CI scenario. It can be observed from the left part of Fig. \ref{fig:achievement}, along with iterations, the gap between the targeted original image and the shape formed by agents gradually diminishes. After 150 iterations, the similarity between the targeted original image and the shape formed by agents exceeds $96.3\%$ and demonstrates the effectiveness of SEAL. Moreover, Fig. \ref{fig:achievement} also gives the impact of the noise of the indirect communication channel on the final performance and shows that the final performance decreases with the increase of standard deviation (e.g., $N_0$) of communication noise, which further reflects the importance of reliable communications on collective intelligence. Meanwhile, Fig. \ref{fig:achievement} provides the comparison between SEAL and other typical algorithms like IQL (independent $\mathcal{Q}$-learning), HQL (hysteretic $\mathcal{Q}$-learning) and LMRL (lenient multi-agent reinforcement learning), and demonstrates that SEAL outperforms IQL, HQL and LMRL in terms of the similarity, thus reflecting the effectiveness of stigmergy within the framework of collective intelligence.

\subsection{The Evaluation of Collective Intelligence}
After demonstrating the effectiveness of CI, we talk about how to evaluate different CI solutions in this part, since agents could collaborate in different organization and communications methods, resulting into distinct intelligence level of CI. In particular, considering the flexibility and universality, we focus on the anytime universal intelligence test (AUIT) in \cite{hernandez-orallo_measuring_2010}, which is information theory-based with a primary focus on the homogeneous agents with the same type of agents and can be easily extended to the case where heterogeneous agents interact in different ways.

\begin{figure}
	\centering
	\includegraphics[width = 0.435\textwidth]{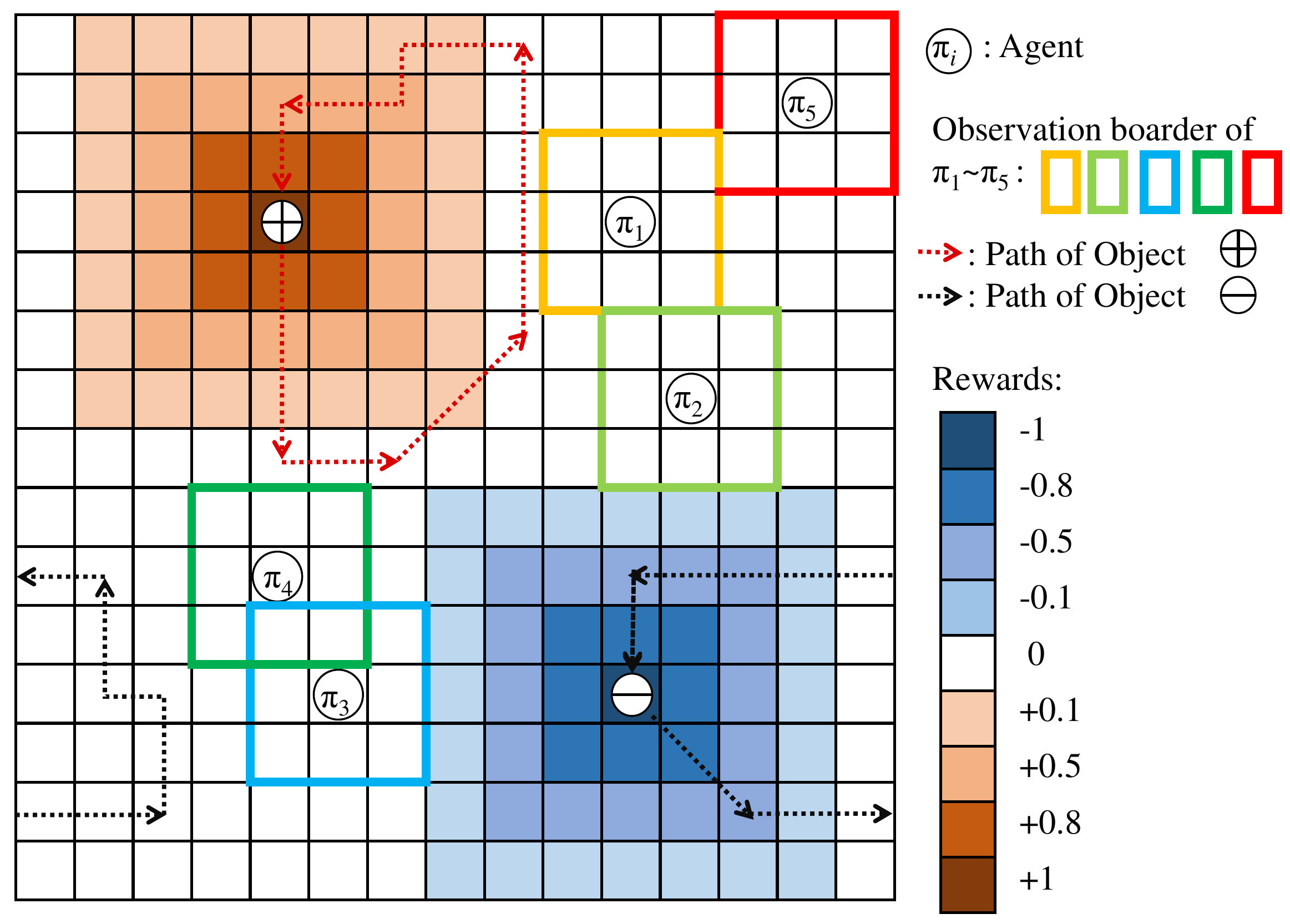}
	\caption{The AUIT model with 5 collective agents and 2 special objects to evaluate the intelligence level of CI.} 
	\label{fig:testmodel}
\end{figure}

AUIT works in a toroidal grid space with periodic boundaries. In other words, an agent, which moves off one border, will appear on the opposite one. 
In AUIT, there are objects from a finite set $\Omega=\left\{  \pi_1, \pi_2, ... , \pi_x, \oplus, \ominus   \right\} $ which contains a set $\Pi$ of collective intelligent agents ($\Pi \in \Omega, \Pi=\left\{ \pi_1, \pi_2, ... , \pi_x\right\} $) and two moving special objects, Good ($\oplus$) and Evil ($\ominus$). The two special objects move in the environment with measurable complexity movement patterns while the movement of agents in $\Pi$ is chosen from a finite set of moving actions $A=\big\{\rm LEFT, RIGHT, UP, DOWN, UP-LEFT, UP-RIGHT, DOWN-LEFT, DOWN-RIGHT, STAY\big\}$. The intelligence level of this CI is determined by a predefined reward function mapped from the distance between the evaluated agents to objects $\oplus$ and $\ominus$ \cite{hernandez-orallo_measuring_2010}. 

As the synergy among the agents is of vital importance to the CI level \cite{liemhetcharat_weighted_2014}, AUIT could be easily extended to different communication scenarios like direct communication (i.e., talking), indirect communication (i.e., stigmergy by digital pheromone), and imitation. In particular, different CI agents in AUIT retrieve information distinctly. For example, agents with direct communication will be informed of all agents' observations exactly, while agents with indirect communication will get others' observations with a specific bias. Besides, agents with imitation only have the capability to get observations from agents within their observation range.

AUIT could easily perform the sensitivity analysis of the CI level. For different task complexities, AUIT could change the two special objects' movement patterns and quantify the complexity in terms of Kolmogorov complexity. Meanwhile, AUIT can differ the search space complexity or environmental complexity by tuning the size of the environment (yielding different Shannon entropies) and remain other settings fixed. 

\section{Conclusion and Future Directions}
In this paper, we have highlighted the CI among connected beings and things, which harnesses the advancement in CT and AI and is assumed to emerge in the 6G era. Following that, we have proposed the IEML as the communicating language among multiple agents to augment human intelligence and also explained the advantages and potential applications of IEML to name information, so as to advance communications. Afterwards, we have demonstrated the achievement by collective intelligent agents through simple indirect communications and have manifested the effectiveness of CI to boost intelligence. In a word, CI would benefit both CT and AI in return.

In fact, towards accomplishing IoI, there still exist many concerns to be addressed in the future.

\begin{itemize}
\item Along with the gradual accumulation of connected intelligent agents, how does the intelligence level vary?
\item What are the reliable realization (e.g., organizational structures) of involved procedures for CI? 
\item Following the preliminary success of AI \cite{xiao_iot_2018}, how does CI potentially affect the security?
\end{itemize} 
We will try to answer these questions in the future.

\section*{Author Biographies}
\textbf{Rongpeng Li} is an assistant professor in Zhejiang University, Hangzhou, China. His research interests currently focus on multi-agent reinforcement learning and network slicing.

\textbf{Zhifeng Zhao} is a Principal Researcher with Zhejiang Lab as well as with Zhejiang University, Hangzhou, China. His research area includes collective intelligence and software-defined networks.

\textbf{Xing Xu} is a Ph.D candidate in Zhejiang University, Hangzhou, China. His research interests  currently focus on collective intelligence.

\textbf{Fei Ni} is a Ph.D candidate in Zhejiang University, Hangzhou, China. Her research interests  currently focus on collective intelligence.

\textbf{Honggang Zhang} is a Professor of Zhejiang University, China. He is currently involved in the research on cognitive green communications.
\end{document}